\documentclass[11pt,letterpaper]{article}
\usepackage{emnlp2016}
\usepackage{times}
\usepackage{latexsym}
\usepackage{adjustbox}
\usepackage{siunitx}
\usepackage{graphicx}
\usepackage{pbox,url}
\usepackage{amsmath,amssymb}
\usepackage{relsize}
\usepackage[small]{caption}
\usepackage{float}

\usepackage{subfig}
\usepackage[T1]{fontenc}
\usepackage{tablefootnote}
\newcommand{\ncell}[2][]{\pbox{6cm}{#1 \\ {\smaller \textit{#2}}}}
\newcommand{\ncellc}[2][]{\parbox{6cm}{ \centering #1 \\ \textsmaller{\textit{#2}} }}
\newcommand{\bst}{\bf}

\DeclareMathOperator{\Tr}{Tr}
\newcommand{\MLP}{\operatorname{MLP}}
\newcommand{\affine}{\operatorname{affine}}

\usepackage{bm}
\usepackage{booktabs}
\usepackage{siunitx}

\usepackage{color}
\newcommand{\model}[2][]{\textsc{#2}$_{\text{#1}}$\xspace}

\newcommand{\word}{\model[word]{bilstm}}
\newcommand{\bilstmchar}{\model[char]{bilstm}}
\newcommand{\bilstmmorph}{\model[morph]{bilstm}}
\newcommand{\avemorph}{\model[morph]{ave}}
\newcommand{\cnnchar}{\model[char]{cnn}}

\usepackage{multirow}
\usepackage{xspace}
\usepackage{multirow}
\usepackage{paralist}

\renewcommand{\vec}[1]{\bm{#1}}

\newcommand{\mat}[1]{\mathbf{#1}}
\newcommand{\vsuper}[2]{\vec{#1}^{(\text{#2})}}
\newcommand{\msuper}[2]{\mat{#1}^{(\text{#2})}}

\newcommand\vs{\vec{s}}
\newcommand\vt{\vec{t}}
\newcommand\va{\vec{a}}

\newcommand\ve{\vec{e}}
\newcommand\vi{\vec{i}}

\newcommand\vh{\vec{h}}
\newcommand\vm{\vec{m}}

\newcommand\vr{\vec{r}}

\newcommand\vhf{\vec{h}^{\rightarrow}}
\newcommand\vhb{\vec{h}^{\leftarrow}}
\newcommand\Rs{\msuper{R}{s}}
\newcommand\rs{\vsuper{r}{s}}
\newcommand\ms{\vsuper{m}{s}}
\newcommand\ws{\vsuper{w}{s}}

\newcommand\Rt{\msuper{R}{t}}
\newcommand\rt{\vsuper{r}{t}}

\newcommand\sizev[1]{\in \mathbb{R}^{#1}}
\newcommand\sizem[2]{\in \mathbb{R}^{#1 \times #2}}

\newcommand\softmax{\operatorname{softmax}}
\emnlpfinalcopy

\title{ Word Representation Models for Morphologically Rich Languages in Neural Machine Translation} 

\newcommand{\spadeaff}{\ensuremath{1}\xspace}
\newcommand{\clubaff}{\ensuremath{2}\xspace}
\author{Ekaterina Vylomova,$^{\spadeaff}$  Trevor Cohn,$^{\spadeaff}$ \and Xuanli He$^{\spadeaff}$ and Gholamreza Haffari$^{\clubaff}$\\ 
  $^{\spadeaff}$Department of Computing and Information Systems, University of Melbourne\\
   $^{\clubaff}$Faculty of Information Technology, Monash University\\
  \texttt{\small evylomova@gmail.com tcohn@unimelb.edu.au} \\ \texttt{\small xuanlih@student.unimelb.edu.au gholamreza.haffari@monash.edu}\\[-0.0ex]}

\date{}

\begin{document}

\maketitle

\begin{abstract}
Dealing with the co
mplex word forms in morphologically rich languages is an open problem in language processing, and is particularly important in translation. 
In contrast to most modern neural systems of translation, which discard the identity for rare words, in this paper we propose several architectures for learning word representations from character and morpheme level word decompositions. 
We incorporate these representations in a novel machine translation model which jointly learns word alignments and translations via a hard attention mechanism. 
Evaluating on translating from several morphologically rich languages into English, we show consistent improvements over strong baseline methods, of between 1 and 1.5 BLEU points.
%In term of the paper, we consider only translation from morphologically rich languages into English. We show that ... outperforms other model achieving ... BLEU score.
\end{abstract}

\section{Introduction}

Models of end-to-end machine translation based on neural networks have been shown to produce excellent translations, rivalling or surpassing traditional statistical machine translation systems \cite{kalchbrenner2013recurrent,sutskever2014sequence,bahdanau2015neural}. 
A central challenge in neural MT is handling rare and uncommon words.
Conventional neural MT models use a fixed modest-size vocabulary, such that 
the identity of rare words are lost, which makes their translation exceedingly difficult.
Accordingly sentences containing rare words tend to be translated much more poorly than those containing only common words \cite{sutskever2014sequence,bahdanau2015neural}.
The rare word problem is particularly exacerbated when translating from 
morphology rich languages, where the several morphological variants of words result 
in a huge  vocabulary with a heavy tail distribution.
For example in Russian, there are at least 70 words for dog, encoding case, gender, age, number, sentiment and other semantic connotations. 
Many of these words share a common lemma, and contain regular morphological affixation; consequently much of the information required for translation is present, but not in an accessible form for models of neural MT.

In this paper, we propose a solution to this problem by constructing word representations compositionally from smaller sub-word units, which occur more frequently than the words themselves. 
We show that these representations are effective in handling  rare words, and 
increase the generalisation capabilities of neural MT beyond the vocabulary observed in the training set. 
We propose several neural architectures for compositional word representations, and systematically compare these methods integrated into a novel neural MT model. 

More specifically, we make use of character sequences or morpheme sequences in building word representations. 
These sub-word units are combined using recurrent neural networks (RNNs), convolutional neural networks (CNNs), or simple bag-of-units.
This work was inspired by research into compositional word approaches proposed for language modelling (e.g., \newcite{botha2014compositional}, \newcite{kim15Jernite}), 
with a few notable exceptions \cite{ling15trancoso,Sennrich15Haddow,Marta2016jose}, these approaches have not been applied to the more challenging problem of translation.
We integrate these word representations into a novel neural MT model to build robust word representations for the source language. %, including rare and unseen words. 
%

%\trevor{This paragraph could lead with the OSM stuff, and end with comparison to Sutsk and Bahd, perhaps?}
%In neural machine translation, translation is considered as sequence to sequence problem %\cite{sutskever2014sequence}, where the source sentence is first encoded in a vector form, from which the %target sentence is generated one word at a time, in a left to right manner. 
%
%This was extended by the \emph{attentional} neural MT \cite{bahdanau2015neural}, where an attention mechanism determines which parts of the source sentence needs to be considered when generating the next word, thus allowing the dynamic use of a much larger source encoding and better matching the cognitive process of human translation.
%
Our novel neural MT model, is based on the operation sequence model (OSM; \newcite{dsf11}, \newcite{fc13}), which considers translation as a sequential decision process.
The decisions involved in generating each target word is decomposed into separate translation and alignment factors, where each factor is modelled separately and conditioned on a rich history of recent translation
decisions.
Our OSM can be considered as a form of attentional encoder-decoder \newcite{bahdanau2015neural} with \emph{hard} attention in which each decision is contextualised by at most one source word, contrasting with the \emph{soft} attention in \newcite{bahdanau2015neural}.
%\reza{the rationale of neural OSM, why integrating word models into OSM?}
%\trevor{could argue that based on vision (Salakutdinov) results for soft vs hard attention, this is a compelling idea; although they don't fix the hard alignments, but learn 'em.} 

Integrating the word models into our neural OSM, we provide -- for the first time -- a comprehensive and 
systematic evaluation of the resulting word 
representations when translating into English from several morphologically rich languages, Russian, Estonian, and Romanian. 
Our evaluation includes both intrinsic and extrinsic metrics, where we compare these approaches 
based  on their translation performance as well as their ability to recover synonyms for the rare words. 
We show that morpheme and character representation of words leads much better heldout perplexity although the improvement on the translation BLEU scores is more modest. 
Intrinsic analysis shows that the recurrent encoder tends to capture more morphosyntactic information about words, whereas convolutional network better encodes the lemma. 
Both these factors provide different strengths as part of a translation model, which might use lemmas to generalise over words sharing translations, and morphosyntax to guide reordering and contextualise subsequent translation decisisions.
These factors are also likely to be important in  other language processing applications.

\section{Related Work}

Most neural models for NLP rely on words as their basic units, and consequently face the problem of how to handle tokens in the test set that are out-of-vocabulary (OOV), i.e., did not appear in the training set (or are considered too rare in the training set to be worth including in the vocabulary.)
Often these words are either assigned a special \texttt{UNK} token,
%, or, in case of MT, acquired from an external vocabulary. \trevor{Kat: last sentence needs justification}
% Not sure about the the
which allows for application to any data, however it comes at the expense of modelling accuracy especially in structured problems like language modelling and translation, where the identity of the word is paramount in making the next decision. 

% There are various classes of OOV words: some of them might be different forms sharing same lemma, therefore could be recovered. 
% Others might present proper nouns that could be either copied or transliterated in case of MT task. And, finally, there is a small percentage of those that present a completely unknown concept, and could hardly be recovered without appropriate definition or context.  
 
%

One solution to OOV problem is modelling sub-word units, using a model of a word from its composite morphemes.
%There is no single model or approach on how the morphemes should be combined to form a word.   
\newcite{luong2013better} proposed a recursive combination of morphs using affine transformation, however this is unable to differentiate between the compositional and non-compositional cases. \newcite{botha2014compositional} aim to address this problem by forming word representations from adding a sum of each word's morpheme embeddings to its word embedding. 
%In the this paper we follow a similar method, in that we combine word embedding with a compositional representation, however we use max pooling for combination instead, and consider both character and morpheme level representations. Moreover we evaluate on translation rather than language modelling. 

%
Morpheme based methods rely on good morphological analysers, however these are only available for a limited set of languages.
Unsupervised analysers  \cite{creutz2007unsupervised} are prone to segmentation errors, particularly on fusional or polysynthetic languages.
%Morphemes are often acquired using unsupervised morphology learning, e.g. using Morfessor CAT-MAP \cite{creutz2007unsupervised}. 
%Therefore, they highly depend on initial settings of the analyser and might yield a word parsing that could be different from the "ground truth". 
In these settings, character-level word representations may be more appropriate.
%
%
%As for the case with morphemes, it is an open question as to which network architecture better predicts a word's meaning from its characters. 
Several authors have proposed convolutional neural networks over character sequences, as part of models of part of speech tagging \cite{icml2014c2_santos14}, language models \cite{kim2015character} and machine translation \cite{Marta2016jose}. 
These models are able to capture not just orthographic similarity, but also some semantics.
Another strand of research has looked at recurrent architectures, using long-short term memory units
\cite{ling2015finding,ballesteros2015improved} which can capture long orthographic patterns in the character sequence, as well as non-compositionality. 

All of the aforementioned models were shown to consistently outperform standard word-embedding approaches. But there is no systematic investigation of the various modelling architectures or comparison of characters versus morpheme as atomic units of word composition. In our work we consider both morpheme and character levels and study 1) wether character-based approaches can outperform morpheme-based, and, importantly, 2) what linguistic lexical aspects are best encoded in each type of architecture, and their efficacy as part of a machine translation model when translating from morphologically rich languages.

\begin{figure}[t]
\centering
\includegraphics[scale=0.2]{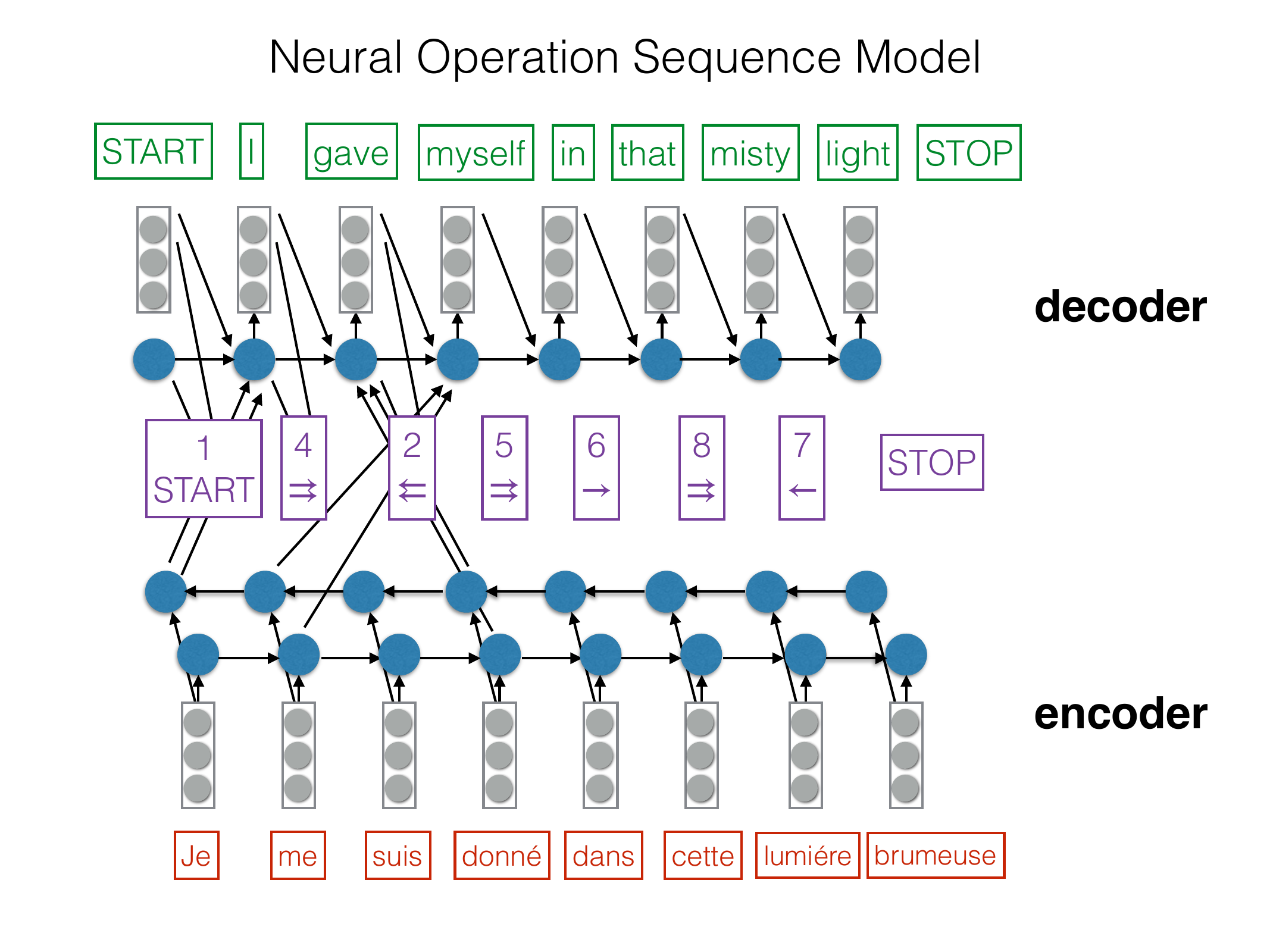}
\caption{Illustration of the neural operation sequence model for an example sentence-pair.  }
\label{fig:osm}
\end{figure}

\section{Operation sequence model}
\label{sec:sentence-osm}
The first  contribution of this paper is a neural network variant of the Operational Sequence Model (OSM) \cite{dsf11,fc13}. 
In OSM,  the translation is modelled as a sequential decision process. 
%In contrast to the original model, our model makes use of word embeddings instead of  words.
%in place of the words the current model learns their corresponding vector embeddings. 
%
The words of the target sentence are generated one at a time
in a left-to right order, similar to the decoding strategy in traditional phrase-based SMT. 
The decisions involved in generating each target word is decomposed into
a number of separate factors, where each factor is modelled separately and
conditioned on a rich history of recent translation decisions.

In previous work \cite{dsf11,fc13}, the sequence of operations is modelled as Markov chain with 
a bounded history, where each translation decision is conditioned on a
finite history of past decisions.
Using deep neural architectures, we model the sequence of translation decisions as a 
non-Markovian chain, i.e. with unbounded history. Therefore, our approach is able to capture long-range
dependencies which are commonplace in translation and missed by previous approaches.

More specifically, the operations are (i) generation of a target word,  
(ii) jumps over the source sentence to capture re-ordering (to allow 
different sentence reordering in the target vs. source language), 
(iii) aligning to NULL to capture gappy phrases, and (iv) finishing the translation process.
The probability of a sequence of operations  to 
generate a target translation $\vt$ for a given source sentence $\vs$ is
$p(\vt,\va|\vs) = $
\begin{align}
\label{train_obj}
 \prod_{j=1}^{|\vt|+1} p(\tau_j | t_{<j}, \tau_{<j}, \vs) \prod_{j=1}^{|\vt|} p(t_j | t_{<j},  \tau_j, \vs) 
%\nonumber
\end{align}
where $\tau_j$ is a jump action moving over the source sentence (to align a target word to a 
source word or null) or finishing the translation process $\tau_{|\vt|+1}=\textrm{FINISH}$.
It is worth noting that the sequence of operations for generating a 
target translation (in a left-to-right order) has a 1-to-1 correspondence to an alignment $\va$, 
so the use of $P(\vt,\va|\vs)$ in the left-hand-side. 

\begin{figure}
\hspace{-4ex}
\includegraphics[width=0.58\textwidth]{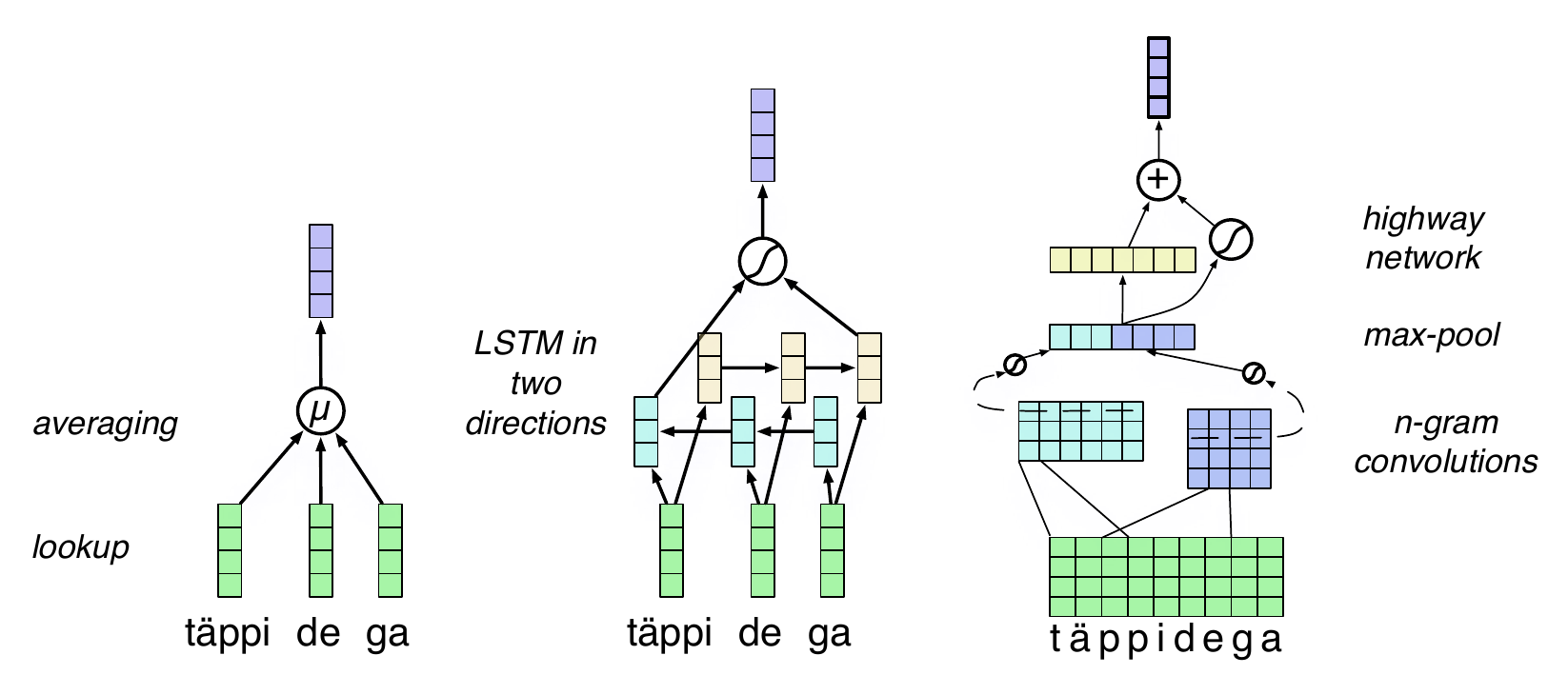}
\caption{Model architecture for the several approaches to learning word representations, showing from left: bag-of-morphs, BiLSTM over morphs, and the character convolution. Note that the BiLSTM is also applied at the character level. The input word, 
\emph{t\"{a}ppi-de-ga}, is Estonian for \emph{speckled}, bearing plural (de) and comitative (ga) suffixes.}
\end{figure}

Our model generates the target sentence and the sequence of operations with a 
recurrent neural network (Figure \ref{fig:osm}). At each stage, the RNN state
is a function of the previous state, the previously generated target word, and 
an aligned source word,
$\vh_{j} = \MLP\left( \vh_{j-1},\rt_{t_{j-1}},\rs_{s_{i_j}} \right) $
%\vh_{j} = \tanh \left( \msuper{W}{h} \vh_{j-1} + \Wti \rt_{t_{j-1}} + \Wsi \rs_{s_{i_j}} + \msuper{b}{h} \right)
using a single layer perceptron (MLP) which applies an affine transformation to the concatentated input vectors followed by a $\tanh$ activation function, where
$\Rt \sizem{V_T}{E_T}$ and $\Rs \sizem{V_S}{E_S}$ are word embedding matrices
with 
%$H$ the number of hidden units,  
$V_S$ the size of the source
vocabulary, $V_T$ the size of the target vocabulary, and $E_T$ and $E_S$ the word embedding 
sizes for the target and source languages, respectively.

The model then generates the target word $t_i$ and index of the  
source word to be translated next,\footnote{The indices 0 and 
$|\vs|+1$ represent the NULL and FINISH operations.}
\begin{eqnarray*}
t_j &\sim  &\softmax   \big( \affine(\vh_j) \big) \\
i_{j+1}  &\sim & \softmax \Big( \Phi(\vs, \vi_{\leq j}, \vt_{\leq j}) \msuper{b}{f}  \\  
& + & \rs  \msuper{W}{sh} \vh_j  + \rs  \msuper{W}{st} \rt_{t_{j}}   \Big) 
\end{eqnarray*}
where $\affine$ performs an affine transformation of its input,\footnote{An affine transform multiplies the input vector by a matrix and adding a bias vector, equivalent to a full connected hidden layer with linear activation.} and the parameters include %$\msuper{W}{to} \sizem{V_T}{H}$, $\msuper{b}{ot} \sizev{V_T}$,
$\msuper{W}{sh} \sizem{E_S}{H}$, $\msuper{W}{ts} \sizem{E_T}{H}$, $\msuper{b}{f} \sizev{F}$, 
and $F$ is the dimensionality of the feature vector $\Phi(.)$ representing the induced alignment structure (explained in the next paragraph). 
The matrix  encoding of the source sentence $\rs \sizem{(|\vs|+2)}{E_S}$ is defined as 
\begin{equation}
\label{input_encoding}
\rs = \left[\vr_{\textrm{NULL}},\rs_{s_1},\ldots,\rs_{s_{|\vs|}}, \vr_{\textrm{FINISH}} \right] 
\end{equation}
where it includes the embeddings of the source sentence words and the NULL and FINISH actions.

The feature matrix $\Phi(|\vs|,\vi_{\leq j}, \vt_{\leq j}) \sizem{(|\vs|+2)}{F}$ captures 
the important aspects between a candidate position for the next alignment
and the current alignment position; this is reminiscent
of the features captured in the HMM alignment model.
The feature vector in each row is composed of two parts :\footnote{More generally, $\Phi(.)$
can capture any aspects of the past alignment decisions, hence it can be used to impose 
structural biases to constrain the alignment space in neural OSM, e.g. symmetry, 
fertility, and position bias.} (i) the first part is
a one-hot vector activating the proper feature  depending whether $i_{j+1}-i_j$ is equal to
$\{ 0, 1, \ge 2, \le -1 \}$ or if the action is NULL or FINISH, and (ii)
the second part consists of two features $i_{j+1}-i_j$ and $\frac{i_{j+1}-i_j}{|\vs|}$.

Note that the neural OSM can be considered as a \emph{hard} attentional model, as opposed to the \emph{soft} attentional neural translation model \cite{bahdanau2015neural}. 
In their soft attentional model, a dynamic summary of the source sentence is used as context to each translation decision, which is formulated as a weighted average of the encoding of all source positions.
In the hard attentional model this context comes from the encoding of a single fixed source position.
This has the benefit of allowing external information to be included into the model, here the predicted alignments from high quality word alignment tools, which have complementary strengths compared to neural network translation models. 

%\reza{maybe to talk about the rationale/motivation?}
%We note that our neural OSM can be considered as a
%\emph{hard} variant of the \emph{attentional} neural translation model \cite{bahdanau2015neural}. 
%In the attnetional models, a \emph{soft} attention
%mechanism is used where the contribution of each source word to the generation of the next target word is 
%avergaed according to an alignment dictribution. In contrast, the alignment distrbution in our model is sharply picked 
%as a point mass distrbution. This leads to a much faster generration of the target sentence, and is particularly 
%useful for those situations where a word in the target sentence is mostly aligned with no more than one word in the source sentence. 
%\reza{maybe need to show in the experiments, are there more rationales?} 

%%% Local Variables: 
%%% mode: latex
%%% TeX-master: "paper2_emnlp2016"
%%% End: 

%\newcommand{\vm}{\mathbf{m}}
\newcommand{\vc}{\mathbf{c}}
\newcommand{\cU}{\mathcal{U}}
\newcommand{\mxpl}{\mathcal{\oslash}}  %max pool notation
\newcommand{\vWuf}{\mathbf{W}_{\rightarrow}^{(uh)}}
\newcommand{\vWhf}{\mathbf{W}_{\rightarrow}^{(hh)}} 
\newcommand{\vbuf}{\mathbf{b}_{\rightarrow}} 
\newcommand{\vWub}{\mathbf{W}_{\leftarrow}^{(uh)}}
\newcommand{\vWhb}{\mathbf{W}_{\leftarrow}^{(hh)}}
\newcommand{\vbub}{\mathbf{b}_{\leftarrow}}

\section{Word Representation Models}

Now we turn to the problem of learning word representations. 
As outlined above, when translating morphologically rich languages, treating word types as unique discrete atoms is highly naive and will compromise translation quality.
For better accuracy, we would need to characterise words by their sub-word units, in order to capture the lemma and morphological affixes, thereby allowing better generalisation between similar word forms.

In order to test this hypothesis, we consider both morpheme and character level encoding methods which we compare to the baseline word embedding approach.
For each type of sub-word encoder we learn two word representations: one estimated from the sub-units and the word embedding.\footnote{We only include word embeddings for common words; rare words share a \texttt{UNK} embedding.}
Then we run max pooling over both embeddings to obtain the word representation, 
$\mathbf{r}_w  = \vm_w ~\mxpl~ \ve_w   $,
where $\vm_w$ is the embedding of word $w$ and $\ve_w$ is the sub-word encoding. The max pooling operation $\mxpl$ captures non-compostionality in the semantic meaning of a word relative to its sub-parts.
We assume that the model would favour unit-based embeddings for rare words and word-based for more common ones.

Let $\cU$ be the vocabulary of sub-word units, i.e., morphemes or characters, $E_u$ be the dimensionality of unit embeddings, and $M \in \mathbf{R}^{E_u \times |\cU|}$ be the matrix of unit embeddings.
Suppose that a word $w$ from the source dictionary is made up of a sequence of units
% \kat{the recursion of U$_w$looked a bit weird, so I replaced it with length of w} 
$\cU_w := [u_1,\ldots,u_{|w|}]$, where $|w|$ stands for the number of constituent units in the word.  
We combine the representation of  sub-word units using a  
LSTM recurrent neural networks (RNN), convolutional neural network (CNN), or simple bag-of-units (described below). 
The resulting word representations are then 
fed to our neural OSM in eqn (\ref{input_encoding}) as the source word embeddings.

\subsection{Bag of Sub-word Units}

This method is inspired by \cite{botha2014compositional} in which 
the embeddings of sub-word units are simply added together,
%, and then max-pooled with the embedding of the word:
%\begin{eqnarray*}
$\ve_w  = \sum_{u \in \cU_w} \vm_{u}$,
%\end{eqnarray*}
where $\vm_u$ is the embedding of sub-word unit $u$.

\subsection{Bidirectional LSTM Encoder}

The encoding of the word  is formulated
using a pair of LSTMs (denoted \emph{bi-LSTM}) one operating
left-to-right over the input sequence and another operating
right-to-left,
$\vhf_j = \text{LSTM}(\vhf_{j-1}, \vm_{u_j})$ 
and $\vhb_j = \text{LSTM}(\vhf_{j+1}, \vm_{u_j}) $
where $\vhf_j$ and $\vhb_j$ are the LSTM hidden states.\footnote{The memory
cells are computed as part of the recurrence, suppressed here for clarity.}
%
%The left-to-right LSTM function is defined as
%\begin{align*}
%\vhf_j = \tanh\left( \vWuf \vm_{u_j} +  \vWhf \vhf_{j-1} +  \vbuf \right)
%\label{eq:sent-rnn-encoder}
%\end{align*}
%where $\vh_0^{\rightarrow} \sizev{H_u}$ is a learned parameter vector, as are
%$\vWuf \sizem{H_u}{E_u}$,  $\vWhf \sizem{H_u}{H_u}$ and $\vbuf \sizev{H_u}$, with
%$H_u$ the number of hidden units.\footnote{Similarly, $\vh_0^{\leftarrow} \sizev{H_u}$, 
%$\vWuf \sizem{H_u}{E_u}$,  $\vWhf \sizem{H_u}{H_u}$ and $\vbuf \sizev{H_u}$ 
%are the parameters of the right-to-left RNN. Note that we use a long short term memory 
%unit \cite{Hochreiter1997} in place of the RNN, shown here for simplicity of exposition.}
%
%\reza{need to fix the notation for the above matrices}
%%\kat{//Why arrows are under the symbol, is it ok?}
% 
The source word is then represented as a pair of hidden states, from left- and right-most states of LSTMs. These are fed into multilayer perception (MLP) with a single hidden layer and a $\tanh$ activation function to form the word representation,
%\begin{equation*} 
$\ve_w =    \MLP \left( \vhf_{|\cU_w|},  \vhb_1 \right)$.
%\end{equation*}

%This encodes not only the word but also its left and right context,
%which can provide important evidence for its translation.

\subsection{Convolutional Encoder}

The last word encoder we consider is a convolutional neural network,
inspired by a similar approach in language modelling \cite{kim15Jernite}.
Let $U_w \sizem{E_u}{|\cU|_w}$ denote the unit-level representation of $w$,
where the $j$th column corresponds to the unit embedding of $u_j$.
The idea of unit-level CNN is to apply a \emph{kernel} $\mathbf{Q}_l \sizem{E_u}{k_l}$
with the width $k_l$ to $U_w$ to obtain a feature map $\mathbf{f_l} \sizev{|\cU|_w - k_l + 1}$.
More formally, for the $j$th element of the feature map the convolutional representation is
$\mathbf{f_l}(j) = \tanh(\langle U_{w,j},\mathbf{Q}_l \rangle + b) $,
where $U_{w,j} \sizem{E_u}{k_l}$ is a slice from $U_w$ which spans the representations of the $j$th unit and its preceding $k_l-1$ units, and $\langle A, B \rangle = \sum_{i,j}A_{ij}B_{ij}= \Tr\left(AB^T\right)$ denotes the Frobenius inner product. For  example, suppose that
the input has size $[4\times 9]$, and a kernel has size $[4\times 3]$ with a sliding step being 1. Then, we obtain a $[1\times7]$ feature map. This process implements a character $n$-gram, where $n$ is equal to the width of the filter. 
The word representation is then derived by max pooling the feature maps of the kernels:
$ \forall l : \quad \mathbf{r}_w(l) = \max_{j} \mathbf{f_l}(j)$.
In order to capture interactions between the character n-grams obtained by the filters, a \emph{highway  network} \cite{srivastava2015training} is applied after the max pooling layer,
%\begin{equation*}
$\ve_w =\vt\odot \MLP(\vr_w)+ (1-\vt) \odot \vr_w $,
%\end{equation*}
where  $t = \MLP_\sigma(\vr_w)$ is a sigmoid gating function which modulates between a $\tanh$ MLP transformation of the input (left component) and preserving the input as is (right component).

\begin{table*}[t]
\centering
\footnotesize
\begin{tabular}{|l|cc|cc|ccc|}
  \hline  \multirow{2}{*}{\textbf{Set}} &
  \multicolumn{2}{c}{\textbf{Train}}  &
  \multicolumn{2}{c}{\textbf{Development}} &
  \multicolumn{3}{c|}{\textbf{Test}}\\
  \cline{2-3}
  \cline{4-5}
  \cline{6-8}
  & tokens & types &  tokens & types &  tokens & types & OOV rate \\
  \hline
Ru-En & 1,639K-1,809K  &  145K-65K & 150K-168K & 35K-18K & 150K-167K     & 35K-18K & 45\% \\
Ro-En & 1,782K-1,806K &  38K-24K & 181K-183K & 13K-9K & 182K-183K &  13K-8K & 30\% \\
Et-En & 1,411K-1,857K &  90K-25K & 141K-188K & 21K-9K & 142K-189K       & 21K-8K & 45\% \\
\hline
\end{tabular}
\caption{Corpus statistics for parallel data between Russian/Romanian/Estonian and English. 
The OOV rate are the fraction of word types in the source language that are in the test set
but are below the frequency cut-off or unseen in training.}
\label{dataset-stats}
\end{table*}

\section{Experiments}

\paragraph{\textbf{The Setup.}} 
We compare the different word representation models based on three morphologically rich languages using 
both exterinsic and intrinsic evaluations.  
For exterinsic evaluation, we investigate their effects in translating to English from 
Estonian, Romanian, and Russian using our neural OSM. 
For intrinsic evaluation, we investigate how accurately the models recover semantically/syntactically related words to a set of given words.

\paragraph{\textbf{Datasets.}}
We use parallel bilingual data from Europarl for Estonian-English and Romanian-English \cite{koehn2005europarl}, and web-crawled parallel data 
for Russian-English \cite{antonova2011building}.
For preprocessing, we tokenize, lower-case, and filter out sentences longer than 30 words.
Furthermore, we apply a frequency threshold of 5, and replace all low-frequency words with a special UNK token.
We split the corpora into three partitions: training (100K), development(10K), and test(10K); Table~\ref{dataset-stats} provides the datasets statistics. 
%

%\begin{table}[t]
%\centering
%\begin{tabular}{l|cc|cc|cc}
%  & words & types & words & types & words & types \\ 
%  \hline
%Ru-En & 1,639K - 1,809K  &  145K - 65K & 150K - 168K & 35K - 18K & 150K - 167K& 35K - 18K\\ 
%Ro-En & 1,782K - 1,806K &  38K - 24K   & 181K - 183K& 13K - 9K   & 182K - 183K&13K - 8K \\ 
%Et-En & 1,411K - 1,857K &  90K - 25K   & 141K - 188K& 21K - 9K   & 142K - 189K& 21K - 8K\\  
%\hline
%\end{tabular}
%\caption{The training/dev/test sets statistics.}
%\label{data-stats}
%\end{table}

\paragraph{\textbf{Morfessor Training.}}
We use Morfessor CAT-MAP \cite{creutz2007unsupervised} to perform morphological analysis needed for morph-based neural models. Morfessor does not rely on any linguistic knowledge, instead it relays on minimum description length principle to 
construct a set of stems, affixes and paradigms that explains the data. 
Each word form is then represented as $(\textrm{prefix})^*(\textrm{stem})^+(\textrm{suffix})^*$.

We ran Morfessor on the entire initial datasets, i.e before filtering out long sentences. 
The word perplexity is the only Morfessor parameter that has to be adjusted. 
The parameter depends on the vocabulary size: larger vocabulary requires higher perplexity number; setting the perplexity threshold to a small value results in over-splitting. 
We experimented with various thresholds and tuned these to yield the most reasonable morpheme inventories.\footnote{
The selected thresholds were 600, 60 and 240 for Russian, Romanian and
Estonian, respectively.}

% Romanian & 394K &  74K &  60 \\ 
% Estonian & 644K &  290K & 240 \\  
% \begin{table}[t]
% \centering
% \footnotesize
% \resizebox{\columnwidth}{!}{%
% \begin{tabular}{lccc}
%   & \textbf{Dataset Size, sentences} &  \textbf{Types} & \textbf{Threshold} \\ 
%   \hline
% Russian & 999K  &  646K & 600	\\ 
% Romanian & 394K &  74K &  60 \\ 
% Estonian & 644K &  290K & 240 \\  
% \hline
% \end{tabular}
% }
% \caption{Morfessor CAT-MAP statistics}
% \label{morph-stats}
% \end{table}

%\begin{table}[t]
%\centering
%\resizebox{\columnwidth}{!}{
%\begin{tabular}{lcc}
%  & \textbf{Development} & \textbf{Test}  \\ 
%  \hline
%\textbf{Ru-En} & 32\% out of 53\% & 32\% out of 53\%	\\ 
%\textbf{Ro-En} & 42\% out of 32\% & 40\% out of 33\% \\ 
%\textbf{Et-En} & 82\% out of 45\% & 83\% out of 46\% \\  
%\end{tabular}
%}
%\caption{Statistics for out-of-vocabulary (OOV) words, and the percentage of them which can be reconstructed %from the morphemes.}
%\label{unknown-stats}
%\end{table}

\begin{table*}
\centering
\footnotesize
%\resizebox{\columnwidth}{!}{%
\begin{tabular}{|l||cc|cc|cc|}
\hline
                Language    &         \multicolumn{2}{c|}{\textbf{Ru-En}} &  \multicolumn{2}{c|}{\textbf{Et-En}}  &  \multicolumn{2}{c|}{\textbf{Ro-En}}  \\
                            &    BLEU & METEOR &  BLEU & METEOR & BLEU & METEOR   \\
\hline \hline
        Phrase-based Baseline              & 15.02          &44.07           & 24.40           &57.23           &  39.68          & 71.25       \\
\hline
        \bilstmchar             & 15.81          & 44.97          &  \textbf{26.14} & 58.47          &  41.10          & 72.13 \\
    \cnnchar             & \textbf{15.94} & \textbf{45.09} &  25.97          & 58.45          &  41.09          & 72.06 \\
       \bilstmmorph             & 15.61          & 44.96          &  \textbf{26.14} & \textbf{58.48} &  \textbf{41.15} & \textbf{72.20} \\
        \word            & 15.70          & 44.98          &  26.03          & 58.33          &  40.97          & 72.15 \\
\hline
\end{tabular}
%}
\caption{BLEU and METEOR scores for re-ranking the test sets. } %(100 Iter/Non-normalized).}
\label{res:bleu}
\end{table*}

\begin{table}
\centering
\resizebox{\columnwidth}{!}{%
\begin{tabular}{|l|cc|cc|cc|}
\hline
Language &
    \multicolumn{2}{c|}{\textbf{Ru-En}} & \multicolumn{2}{c|}{\textbf{Et-En}} & \multicolumn{2}{c|}{\textbf{Ro-En}} \\
\hline
  & W & A & W & A & W & A \\
\hline
\word & 15.81 & 5.95 & 6.71 & 4.93  &  3.35 & 3.22 \\
\bilstmchar & 14.64  & 5.22  & 5.75  & 4.62 & 3.37  & 2.95 \\
\cnnchar & \bst 13.02 &\bst 5.10 &\bst 5.44 & 4.52  &\bst  3.20 &\bst 2.95\\
\avemorph & 15.96 & 5.12 & 5.91 & \bst 4.48  & 3.30 & 3.13 \\
\bilstmmorph & 15.89 & 5.19 &5.61 & 4.63 & 3.30 & 3.14\\
\hline
\end{tabular}
}
\caption{Word (W) and alignment (A) perplexities results for the development data.}
\label{res:pplx}
\end{table}

Table \ref{dataset-stats} presents the percentage of unknown words in the test for each source language . For reconstruction we considered the words from the native alphabet only. The recovering rate depends on the model. For characters all the words could be easily rebuilt. In case of morpheme-based approach the quality mainly depends on the Morfessor output and the level of word segmentation. In terms of morphemes, Estonian presents the highest reconstruction rate, therefore we expect it to benefit the most from the morpheme-based models. Romanian, on the other hand, presents the lowest unknown words rate being the most morphologically simple out of the three languages. Morfessor quality for Russian was the worst one, so we expect that Russian should mainly benefit from character-based models.

\subsection{Extrinsic Evaluation: MT}

\paragraph{\textbf{Training.}}
We annotate the  training sentence-pairs with their sequence of operations to training the neural OSM model.
We first run a word aligner\footnote{We made use of {\bf fast\_align} in our experiments
\url{https://github.com/clab/fast_align}.} to align each target word to a source word. 
We then read off the sequence of operations by scanning the target words in a left-to-right order. 
As a result, the training objective consists of maximising the joint probability of
target words and their alignments eqn \ref{train_obj}, which is performed by stochastic gradient descent (SGD). 
The training stops when the likelihood objective on the development set starts decreasing.

For the re-ranker, we use the standard features generated by moses\footnote{\url{https://github.com/moses-smt}.} 
as the underlying phrase-based MT system plus two additional features coming from 
the neural MT model. The neural features are based on the generated alignment and the translation probabilities, 
which correspond to  the first and second terms in eqn \ref{train_obj}, respectively.  
We train the re-ranker using MERT \cite{och2003minimum} with 100 restarts.

\paragraph{\textbf{Translation Metrics.}} We use BLEU \cite{papineni2002bleu} and METEOR\footnote{\url{http://www.cs.cmu.edu/~alavie/METEOR/}.} \cite{denkowski2014meteor}
to measure the translation quality against the reference. 
BLEU is purely based on the exact match of $n$-grams  in the generated and reference translation, whereas METEOR 
takes into account matches based on stem, synonym, and paraphrases as well. This is particularly suitable for our 
morphology representation learning methods since they may result in using the translation of paraphrases.  
We train the paraphrase table of METEOR using the entire initial bilingual corpora based on pivoting \cite{bannard2005paraphrasing}.

\paragraph{\textbf{Results.}} 
Table \ref{res:pplx} shows the translation and alignment perplexities of the development sets when the models are trained.
As seen, the \cnnchar model leads to lower word and alignment perplexities in almost 
all cases. This is interesting, and shows the power of this  model in fitting to  morphologically complex languages using only their characters. 
Table \ref{res:bleu} presents BLEU and METEOR score results, where the re-ranker is optimised by the METEOR and BLEU
when reporting the corresponding score. 
As seen, re-ranking based on neural models' scores outperforms the phrase-based baseline. 
Furthermore, the translation quality of the \bilstmmorph model outperforms others for Romanian and Estonian, 
whereas the \cnnchar model outperforms others for Russian which is consistent with our expectations. We assume that replacing Morfessor with real morphology analyser for each language should improve the performance of morpheme-based models, but leave it for future research.
However, the translation quality of the neural models are not significantly different, which  may be due to
the convoluted contributions of high and low frequency words into BLEU and METEOR. 
Therefore, we investigate our representation learning models intrinsically in the next section.

%\begin{table}[!htb]
%\centering
%\resizebox{\columnwidth}{!}{%
%\begin{tabular}{|l|ccc|}
%\hline
%Language                & \textbf{Ru-En} & \textbf{Et-En} & \textbf{Ro-En} \\
%\hline \hline
%Phrase-based Baseline   & 15.02 & 24.40 &  39.68\\
%\hline
%\word          & 15.87 & 26.03 &  40.97\\ 
%\bilstmchar        & 15.89 & 26.14 &  41.10 \\
%\cnnchar &  15.94 &  25.97 & 41.09 \\
%\bilstmmorph            & 15.68 & 26.14 & 41.15\\
%\hline
%Oracle & 21.04 & 33.25 & 47.96\\
%\hline
%\end{tabular}
%}
%\caption{BLEU score results for test set (100 Iter/Non-normalized).}
%\label{res:bleu}
%\end{table}

\subsection{Intrinsic Evaluation}

We now take a closer look at the embeddings learned by the models, based on how well they capture the 
\textit{semantic} and \textit{morphological} information in the  nearest neighbour words. 
Learning representations for low frequency words is harder than that for high-frequency words,
since they cannot capitalise as reliably on their contexts. 
Therefore, we split the test lexicon into 6 subsets according to their frequency in the training set: 
[0-4], [5-9], [10-14], [15-19], [20-50], and 50+. Since we set out word  frequency threshold to 5 for the training set, 
all words appearing in the frequency band [0,4] are in fact OOVs for the test set. 
For each word of the test set, 
we take its top-20 nearest neighbours from the whole training lexicon (without threshold) using cosine metric. 
%

%Therefore, we are aiming at evaluating various word representations depending on the architecture(bilstm vs convolutional), units types, and how word frequency affects the quality of the embedding. 
%As a similarity measure we consider cosine value.  To avoid noise, we additionally apply a threshold of 0.5, and consider only those that are above the value.

\paragraph{\textbf{Semantic Evaluation.}} We investigate how well the nearest neighbours are                   
interchangable with a query word in the translation process.    
So we formalise the notion of semantics of the source words based on their translations in the target language.      
We use \textit{pivoting} to define the probability of a candidate word
$e'$ to be the synonym of the query word $e$, $p(e'|e) = \sum_{f} p(f|e) p(e'|f)$,
where $f$ is a target language word, and the translation probabilities inside the summation are estimated using a word-based    
translation model trained on the entire bilingual corpora (i.e. before splitting into train/dev/test sets).
We then take the top-5 most probable words as the gold synonyms for each query word of the test set.\footnote{We remove query words whose
frequency is less than a threshold in the initial bilingual corpora, since pivoting may not result in high quality synonyms for such words.}   

We measure the quality of predicted nearest neighbours using the multi-label accuracy,\footnote{We evaluated using mean reciprocal rank (MRR) 
measure as well, and obtained results consistent with the multi-label accuracy (omitted due to space constraints).} 
%\begin{equation}
%\label{mult-acur}
$ \frac{1}{|S|} \sum_{w \in S} \mathbf{1}_{[\textrm{G}(w) \cap \textrm{N}(w) \not= \varnothing}]$,
%\end{equation}
where $G(w)$ and $N(w)$ are the sets of gold standard synonyms and nearest neighbors for $w$ respectively; the function $\mathbf{1}_{[C]}$ 
is one if the condition $C$ is true, and zero otherwise. In other words, it is the fraction of words in $S$ 
whose nearest neighbours and gold standard synonyms have non-empty overlap. 

Table \ref{res:NN-sem} presents the semantic evaluation results. As seen, on words with frequency $\leq 50$, 
the \cnnchar model performs best across all of the three languages. Its superiority is particularly interesting for 
the OOV words (i.e. the frequency band [0,4]) where the model has cooked up the representations completely based on the 
characters. For high frequency words (> 50), the \word outperforms the other models.

%For our analysis we introduce binary precision $Pb$ that produces a boolean value if there is any intersection between the gold standard and the model's list of nearest neighbours. At the end, we average these values across the whole set of words in each subset.  

\begin{table}
\centering
%\footnotesize
\resizebox{\columnwidth}{!}{
\begin{tabular}{|l|c|c|c|c|c|c|c|}
\hline
 \textbf{Model \ Freq.}     & \textbf{0-4} & \textbf{5-9} & \textbf{10-14} & \textbf{15-19} & \textbf{20-50} & \textbf{50+}\\
\cline{1-7}
    \multicolumn{7}{|c|}{\bf Russian} \\
\cline{1-7}
\hline
\word        & -          & 0.36  & 0.49 & 0.61 & 0.76 &\bf 0.91  \\
\bilstmchar  & 0.16  & 0.34  & 0.48  & 0.59 & 0.71 & 0.85  \\
\cnnchar     & \bf 0.43  &\bf 0.71  &\bf  0.77  &\bf 0.77 &\bf 0.81 & 0.81  \\
\avemorph    & 
0.03  & 0.21  & 0.33  & 0.40  & 0.55 & 0.78  \\
\bilstmmorph & 0.01  & 0.24  & 0.38 & 0.49 & 0.65 & 0.85 \\
\cline{1-7}
    \multicolumn{7}{|c|}{\bf Romanian} \\
\cline{1-7}
\hline
\word        & -          & 0.47  & 0.63 & 0.71  & 0.81  &\bf 0.91 \\
\bilstmchar  & 0.02  & 0.41  & 0.55 & 0.62    & 0.74 & 0.83 \\
\cnnchar     &\bf 0.59  &\bf 0.82  &\bf 0.81 &\bf 0.84  &\bf 0.88 &  0.84\\
\avemorph     & 0.05  & 0.40  & 0.52 & 0.61 & 0.71 & 0.84  \\
\bilstmmorph  & 0.01  & 0.38  & 0.53 & 0.61 & 0.72 & 0.82 \\
\cline{1-7}
    \multicolumn{7}{|c|}{\bf Estonian} \\
\cline{1-7}
\hline
\word    &  -             & 0.48  & 0.62 & 0.70 &\bf 0.79  &\bf 0.90 \\
\bilstmchar  & 0.13  & 0.39  & 0.48 & 0.55 & 0.63 &  0.78 \\
\cnnchar     &\bf 0.48 &\bf 0.70 &\bf 0.75 &\bf 0.76 & 0.78 & 0.78\\
\avemorph    & 0.07  & 0.29  & 0.40 & 0.47 & 0.56 & 0.76 \\
\bilstmmorph  & 0.013  & 0.36  & 0.45 & 0.52 & 0.60 & 0.76 \\
\hline
\end{tabular}
}
\caption{Semantic evaluation of nearest neighbours using multi-label accuracy on words in different frequency bands.}
\label{res:NN-sem}
\end{table}

\paragraph{\textbf{Morphological Evaluation.}} We now turn to evaluating  the morphological component. 
For this evaluation, we focus on Russian since it has a notoriously hard morphology.
We run another morphological analyser, \textit{mystem} \cite{segalovich2003fast}, to generate \textit{linguistically tagged} morphological analyses for a word, e.g.
POS tags, case, person, plurality, etc. 
We represent each morphological analysis with a bit vector showing the presence of these grammatical features.
Each word is then assigned a set of bit vectors corresponding to the set of its morphological analyses.  
As the \emph{morphology similarity} between two words, we take the minimum of Hamming similarity\footnote{The Hamming similarity is the number of bits having the same value in two given bit vectors.} between the corresponding two sets of bit vectors.
Table \ref{res:NN-sem-tags-lemmas}(a) shows the average morphology similarity between the words and their nearest neighbours across the frequency bands. 
Likewise, we represent the words based on their lemma features; Table \ref{res:NN-sem-tags-lemmas}(b) shows the average lemma similarity.
We can see that both character-based models capture morphology far better than morpheme-based ones, especially in the cases of OOV words. But it is also clear that CNN tends to outperform bi-LSTM in case where we compare lemmas, and bi-LSTM seems to be better at capturing affixes.

\begin{table}
\centering
\resizebox{\columnwidth}{!}{%
\begin{tabular}{c}
%%%%
\begin{tabular}{|l|c|c|c|c|c|c|c|}
\hline
\textbf{Model \textbackslash~Freq.} & \textbf{0-4} & \textbf{5-9} & \textbf{10-14} & \textbf{15-19} & \textbf{20-50} & \textbf{50+}\\
\cline{1-7}
    \multicolumn{7}{|c|}{\bf Russian} \\
\cline{1-7}
\hline
\word         & -    & 0.74 & 0.77 & 0.78 & 0.81 & 0.84 \\
\bilstmchar   &\bst 0.91 &\bst 0.84 &\bst 0.85 &\bst 0.85 &\bst 0.85 &\bst 0.86\\
\cnnchar      & 0.79 & 0.80 & 0.79 & 0.79 & 0.79 & 0.79\\
\avemorph     & 0.74 & 0.73 & 0.76 & 0.77 & 0.79 & 0.79\\
\bilstmmorph  & 0.72 & 0.75 & 0.77 & 0.79 & 0.80 & 0.83\\
\hline
\end{tabular} \\
(a) \\
%%%%
\begin{tabular}{|l|c|c|c|c|c|c|c|}
\hline
\textbf{Model \textbackslash~Freq.} & \bf 0-4 & \bf 5-9 &\bf 10-14 &\bf 15-19 &\bf 20-50 &\bf 50+\\
\cline{1-7}
    \multicolumn{7}{|c|}{\bf Russian} \\
\cline{1-7}
\hline
\word        & -     & 0.03 & 0.05 & 0.06 & 0.09 & 0.15 \\
\bilstmchar  & 0.05  & 0.05 & 0.08 & 0.10 & 0.13 & 0.18\\
\cnnchar     & \bst 0.20  &\bst 0.37 &\bst 0.41 &\bst 0.42 &\bst 0.44 &\bst 0.41\\
\avemorph    & 0.02  & 0.02 & 0.03 & 0.04 & 0.06 & 0.12\\
\bilstmmorph & 0.00  & 0.02 & 0.04 & 0.06 & 0.09 & 0.15\\
\hline
\end{tabular} \\
(b)
%%%%%
\end{tabular}
}
\caption{Morphology analysis for nearest neighbours based on (a) Grammar tag features, and (b) Lemma features.}
\label{res:NN-sem-tags-lemmas}
\end{table}

Now we take a closer look at the character-based models. We manually created a set of non-existing Russian words of three types. Words in the first set consist of known root and affixes, but their combination is atypical, although one might guess the meaning. The second type corresponds to the words with non-existing(nonsense) root, but meaningful affixes, so one might guess its part of speech and some other properties, e.g. gender, plurality, case. Finally, a third type comprises of the words with all known root and morphemes, but the combination is absolutely not possible in the language and the meaning is hard to guess.

Table \ref{res:NN-example} shows that CNN is strongly biased towards longest substring matching from the beginning of the word, and it yields better recall in retrieving words sharing same lemma. Bi-LSTM, on the other hand, is mainly focused on matching the patterns from both ends regardless the middle of the word. And it results in higher recall of the words sharing same grammar features.   

\begin{table}[!h]
\centering
\footnotesize 
\resizebox{\columnwidth}{!}{  
\begin{tabular}{|l|l|}
\multicolumn{2}{c}{\ncellc[\textbf{\textit{Nemec}-kos\'t}]{\textit{German}-ness (s,f,nom,sg)}} \\[2ex] 
%\multicolumn{2}{c}{\ncellc[\textbf{\textit{Derev}-jan-n-ee}]{\textit{Wood}-en-er (a,comp)}}\\[2ex]
\hline
\cnnchar & \bilstmchar \\
\hline
\ncell[nemec+k+oe]{german (a,nom,sg,plen,n)}  & \ncell[ne+blagodar+n+os\'t]{ingratitude (s,f,nom,sg)} \\ 
\hline
\ncell[nemec+k+om]{german (a,abl,sg,plen,n)} & \ncell[ne+forma\'l +n+os\'t]{informality (s,f,nom,sg)} \\ 
\hline
\ncell[nemec+k+ogo]{german (a,gen,sg,plen,n)}  & \ncell[ne+mysl+im+os\'t]{unthinkableness (s,f,nom,sg)} \\ 
\hline
\ncell[nemec+k+oi]{german (a,gen,sg,plen,f)} & \ncell[ne+kompeten+t+n+os\'t]{incompetence (s,f,nom,sg)} \\ 
\hline
\ncell[nemec]{german (s,m,anim,nom,sg)} & \ncell[ne+gramot+n+os\'t]{illiteracy (s,f,nom,sg)} \\ \hline 
\multicolumn{2}{c}{}  \\[-1.5ex]
\multicolumn{2}{c}{\ncellc[\textbf{\textit{Butjav}-ka}]{\textit{Butjav}-vy (s,f,sg,nom,nons)}}\\[2ex]
\hline
 \cnnchar & \bilstmchar \\
\hline
\ncell[bu\'l var+a]{boulvard (s,m,gen,sg)} & \ncell[dubin+k+a]{truncheon (s,f,nom,sg)} \\
\hline
\ncell[bu\'l var+e]{boulvard (s,m,abl,sg)} & \ncell[d\v zorak+a]{djoraka (s,f,nom,sg,nons)}\\
\hline
\ncell[bu\'l var]{boulvard (s,m,nom,sg)} &  \ncell[\v sirot+a]{latitude (s,f,nom,sg)} \\ 
\hline
\ncell[bulav+ki]{pins (s,f,nom,pl)} & \ncell[mozambik]{mozambique (s,geo,m,nom,sg)}\\ 
\hline
\ncell[bu\'l var+ov]{boulvard (s,m,gen,pl)} & \ncell[mazeka]{mazeka (s,f,nom,sg,nons)} \\
\hline
\multicolumn{2}{c}{}  \\[-1.5ex]
\multicolumn{2}{c}{\ncellc[\textbf{Pere-\textit{ton}-ul-sja}]{Re-\textit{sunk} himself (v,pf,intr,praet,sg,indic,m)}}\\[2ex]
\hline
 \cnnchar & \bilstmchar \\
\hline
\ncell[pere+me\v s+a+l+i+\'s]{mixed (v,pf,intr,praet,pl,indic)} & \ncell[pere+mest+i+l+sja]{moved (v,intr,praet,sg,indic,m,pf)}\\ 
\hline
\ncell[pere+rug+a+l+o+\'s]{squabbled (v,pf,intr,praet,sg,indic,n)} & \ncell[za+pis+a+l+sja]{enrolled (v,intr,praet,sg,indic,m,pf)} \\
\hline
\ncell[pri+kos+n+u+l+sja]{touched (v,intr,praet,sg,indic,m,pf)} & \ncell[za+blud+i+l+sja]{lost (v,pf,intr,praet,sg,indic,m)}\\ 
\hline
\ncell[pri+kas+n+u+l+sja]{touched (v,intr,praet,sg,indic,m,pf,inc)} & \ncell[za+pusk+a+l+sja]{launched (v,ipf,intr,praet,sg,indic,m)}\\ 
\hline
\ncell[pere+rod+i+l+sja]{reborn (v,intr,praet,sg,indic,m,pf)} & \ncell[pere+men+i+l+a+s]{changed (v,pf,intr,praet,sg,indic,f)} \\
  \hline

\end{tabular}
}
\caption{\scriptsize {Analysis of the five most similar words to nonsense Russian words, under the \cnnchar and \bilstmchar word encodings based on cosine similarity. The diacritic \'{ } indicates softness.
\textbf{POS  tags}: \emph{s}-noun, \emph{a}-adjective, \emph{v}-verb; \textbf{Gender}: \emph{m}-masculine, \emph{f}-feminine, \emph{n}-neuter; \textbf{Number}: \emph{sg}-singular, \emph{pl}-plural; \textbf{Case}: \emph{nom}-nominative, \emph{gen}-genitive, \emph{dat}-dative, \emph{acc}-accusative, \emph{ins}-instrumental, \emph{abl}-prepositional, \emph{loc}-locative; \textbf{Tense}: \emph{praes}-present, \emph{inpraes}-continuous, \emph{praet}-past, \emph{pf}-perfect, \emph{ipf}-imperfect; \emph{indic}-indicative; \textbf{Transitivity}: \emph{trans}-transitive, \emph{intr}-intransitive; \textbf{Adjective form}: \emph{br}-brevity, \emph{plen}-full form, \emph{poss}-possessive; \textbf{Comparative}: \emph{supr}-superlative, \emph{comp}-comparative; \textbf{Noun person}: \emph{1p}-first, \emph{2p}-second, \emph{3p}-third;\textbf{Other}: \emph{geo}-geolocation, \emph{nons}-nonsense, \emph{inc}-incorrect spelling, \emph{famn}-family name, \emph{praed}-predicative}}
\label{res:NN-example}
\end{table}

%%% Local Variables: 
%%% mode: latex
%%% TeX-master: "paper2_emnlp2016"
%%% End: 

\section{Conclusion}

This paper proposes a novel translation model incorporating a hard attentional mechanism. 
In this context, we have compared four different models of morpheme- and character-level word representations for the source language. 
These models lead to more robust encodings of words in morphological rich languages, and overall better translations than simple word embeddings. 
Our detailed analyses have shown that word-embeddings are superior for frequent words, 
whereas convolutional method is best for handling rare words. 
Comparison of the convolutional and recurrent methods over character sequences has shown that 
the convolutional method better captures the lemma, which is of critical importance for translating out-of-vocabulary words, 
and would also be key in many other semantic applications.
%%% Local Variables: 
%%% mode: latex
%%% TeX-master: "paper2_naaclhlt2016"
%%% End: 

\bibliography{cite-strings,sentence,cite-definitions}
\bibliographystyle{naaclhlt2016}

\end{document}